\documentclass[letterpaper, 10 pt, conference]{ieeeconf}  

\IEEEoverridecommandlockouts                              

\usepackage{formatting} 

\title{\LARGE \bf
Inter-finger Small Object Manipulation with DenseTact Optical Tactile Sensor
}

\author{Won Kyung Do, Bianca Aumann, Camille Chungyoun, and Monroe Kennedy III 
\thanks{Authors are members of the ARMLab in the Mechanical Engineering Department, Stanford University, Stanford, CA 94305, USA. 
{\tt\small \{wkdo, biancalj, camillec, monroek\}@stanford.edu.} 
The first author is supported by a fellowship from the Kwanjeong Educational Foundation. This work is supported by the National Science Foundation under Grants 2142773 and 2220867. Project website with videos are available here: \href{https://sites.google.com/view/inter-finger-manipulation}{https://sites.google.com/view/inter-finger-manipulation} } }

\begin{document}

\maketitle
\thispagestyle{empty}
\pagestyle{empty}

\begin{abstract}

The ability to grasp and manipulate small objects in cluttered environments remains a significant challenge. This paper introduces a novel approach that utilizes a tactile sensor-equipped gripper with eight degrees of freedom to overcome these limitations. We employ DenseTact 2.0 for the gripper, enabling precise control and improved grasp success rates, particularly for small objects ranging from 5mm to 25mm. Our integrated strategy incorporates the robot arm, gripper, and sensor to manipulate and orient small objects for subsequent classification effectively. We contribute a specialized dataset designed for classifying these objects based on tactile sensor output and a new control algorithm for in-hand orientation tasks. Our system demonstrates 88\% of successful grasp and successfully classified small objects in cluttered scenarios. 
\end{abstract}

\section{INTRODUCTION}

Grasping objects commonly found in daily environments is essential for human-robot collaboration tasks. Nevertheless, in-hand manipulation and grasping in cluttered settings continue to pose significant challenges in robotics. Recent research has increasingly focused on incorporating tactile feedback as a vital element in control systems to manage contact kinematics and manipulation tasks more effectively.

Despite this, the specific issue of grasping small objects in cluttered environments remains largely unresolved. When a robot interacts with an object, the situation changes, requiring a revised approach. This adaptability is common in human interactions but challenging for robots. The solution involves enabling robots to manipulate or identify small objects in cluttered scenarios.

Tactile sensors are instrumental in overcoming these issues. When grasping objects in cluttered spaces, traditional external vision systems often prove insufficient. Visuotactile sensors, however, offer a remedy by providing high-resolution data in localized areas. Additionally, hemispherical tactile sensors like DenseTact offer enhanced sensing capabilities and greater adaptability in terms of deformation, which is advantageous for compliance control.

In this study, we use tactile sensing and extra degrees of freedom on the gripper to tackle grasping, manipulating, and classifying small objects in cluttered environments. The transient dynamics of small objects, simulation challenges, and inadequacy of traditional controls post-grasp complicate the problem.

\begin{figure}[t]
\begin{center}
\centerline{\includegraphics[width=\columnwidth]{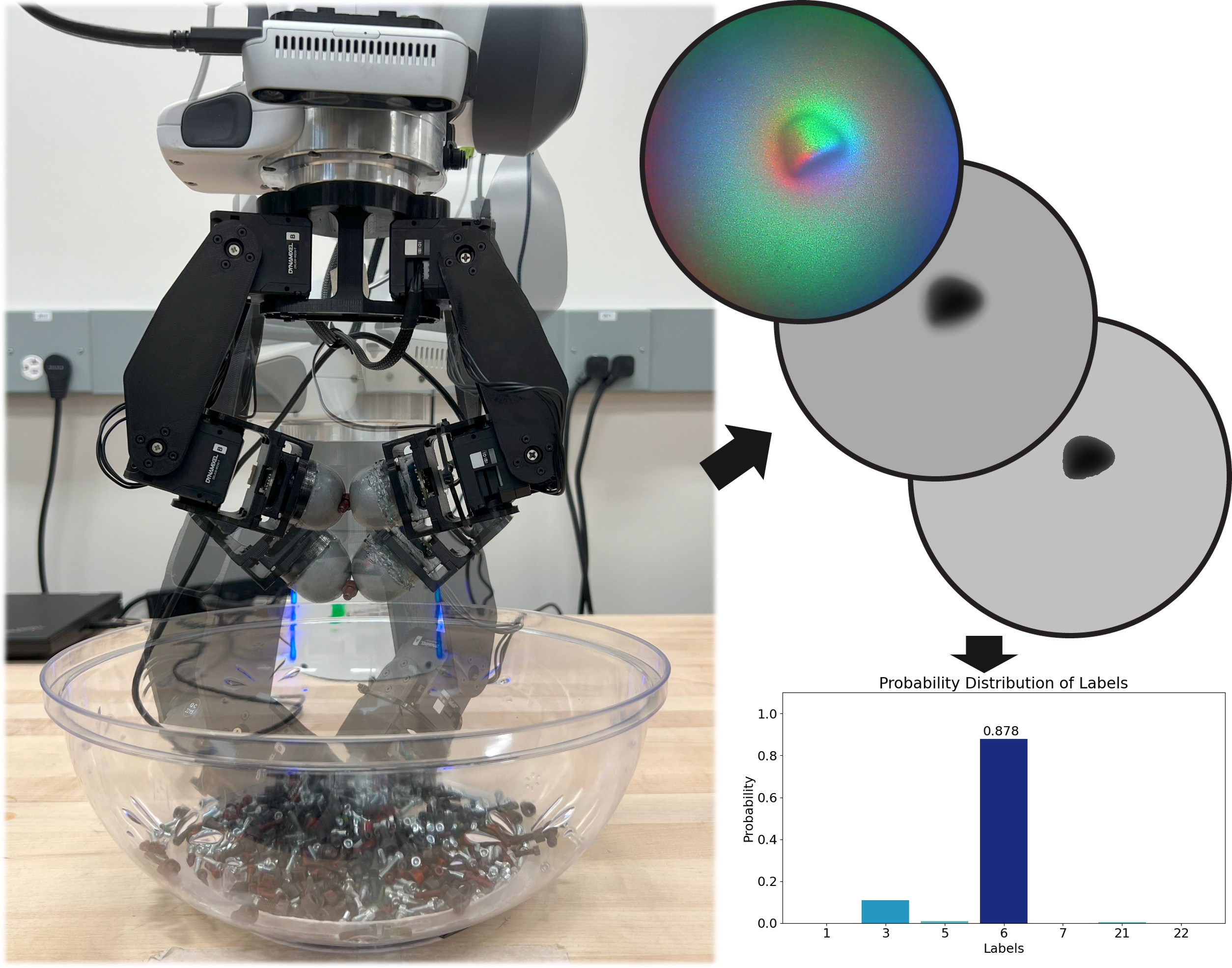}}
\caption{\textbf{Overview of the grasping and classifying of small objects in cluttered environments.} The left image shows the process of grasp and control to classify the object, the right top shows the result of images from the sensor, and the right bottom shows the result of classification.}
\label{mainfig}
\end{center}
\vskip -0.4in
\end{figure}
The primary contributions of this paper are:

\begin{enumerate}
\item Development of a novel gripper with DenseTact 2.0, featuring 8 degrees of freedom for rolling manipulation.
\item Establishment of an integrated strategy involving the robot arm, gripper, and sensor for the manipulation and orientation of small objects for classification.
\item Creation of a dataset for classifying small objects based on tactile sensor outputs.
\item Successful classification and manipulation of objects smaller than the sensor and gripper sizes.
\item Design of a new control algorithm for in-hand orientation tasks involving `unknown' small objects.
\end{enumerate}

The paper is structured as follows: Section \ref{sec:related} reviews related works; Section \ref{sec:probstatement} outlines the problems addressed; Section \ref{sec:method} discusses the methodologies for gripper development, perception, object grasping, manipulation, and classification; Section \ref{sec:experiment} presents the results and demonstrations, and conclusions and future work are discussed in Section \ref{sec:conclusion}.

\begin{figure*}[t]
\vskip 0.1in
\begin{center}
\centerline{\includegraphics[width=0.98\textwidth]{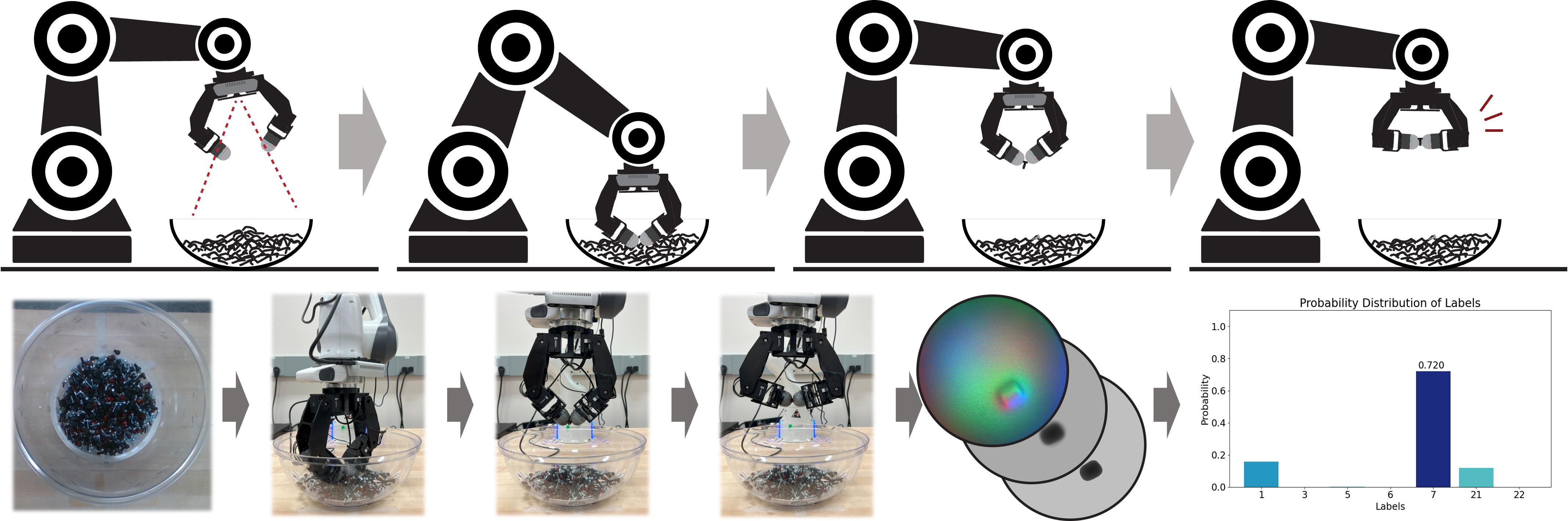}}
\caption{Pipeline of the small-object grasping, pinching, classifying, and sorting process.}
\label{pic:pipeline}
\end{center}
\vskip -0.4in
\end{figure*}
\section{Related Works} \label{sec:related}

Grasping and manipulating small objects through tactile sensor input is a complex endeavor. A plethora of research initiatives have been aimed at various facets of this task. Specifically, in-hand manipulation has emerged as an active research domain in recent years. Works such as those cited in \cite{teeple2022controlling, chavan2018hand} have proficiently tackled challenges associated with continuous contact or variations in friction during object interactions. In-hand manipulation employing external vision is discussed in \cite{handa2022dextreme, morgan2022complex} and the use of Adaptive RL  (reinforcement learning) policy derived from simulation torque input in robotic hands \cite{qi2022hand}. However, despite solutions to the sim-to-real problem, grasping objects in cluttered environments complicates policy training. Both model-based and model-free RL approaches often struggle in such dynamically altering environments. Moreover, relying solely on external vision for object orientation may become unfeasible during gripping, as the object becomes partially or fully occluded.

Tactile sensors, particularly visual tactile sensors, play a crucial role in in-hand manipulation and classification tasks. Placing a tactile sensor at the tip of the gripper enables intricate activities such as cable manipulation \cite{she2021cable}, box packing \cite{dong2019tactile}, 3D pinching between fingers \cite{psomopoulou2021robust, mao2023learning}, and grasping of both soft and rigid objects \cite{welle2023enabling}. However, these tasks primarily focus on manipulating larger objects or involve specialized object manipulation, thus limiting their generalizability for handling small objects.

Tactile sensors are also effective in object or environmental classification. They can detect the hardness of objects, whether the sensor is vision-based or electrical transduction-based \cite{solano2023embedded, yuan2016estimating, Andrussow2023Minsight, DRIMUS20143}. Classification of objects can be accomplished using multiple tactile sensors in a single grasp \cite{9732681} or with vision-based tactile sensors \cite{9967308}. However, the majority of these sensors are designed for classifying larger or deformable objects and may not be appropriate for small object classification in cluttered environments due to issues such as sensing resolution and gripper size. To address these challenges, we have developed a new gripper equipped with a sensor designed to both manipulate and classify small objects from a single grasp.


\section{Problem Statement} \label{sec:probstatement}

This paper addresses the integrated tasks of grasping, reorienting, and classifying small objects ($5mm \sim 25mm$) using optical tactile sensor input, all in a quasi-static state. The objects are smaller than the sensor size (30mm diameter). The components of the problem statement in this paper are defined as follows:
\begin{itemize}
    \item \textbf{Grasping in Cluttered environment}: The primary challenge is grasping a small object from a cluttered bowl. We assume that the gripper interacts only with the objects, not the bowl itself. The task is solved using a robotic arm equipped with soft tactile sensors.
    \item \textbf{Object Reorientation}: After it is grasped, the object must be reoriented within the gripper for stable holding. This is achieved using a multi-degree of freedom (DOF) gripper.
    \item \textbf{Object Classification}: Finally, classification is performed using the tactile sensor on the gripper. Vision-based methods are unsuitable due to occlusion when the object is grasped.
\end{itemize}

\section{Method} \label{sec:method}

\subsection{Hardware setup}

\subsubsection{Gripper for inter-finger manipulation}

Numerous gripper designs have been proposed for various tasks \cite{zhang2020state}. Among these, grippers capable of grasping small objects in cluttered environments often focus on specific usage or are limited to two parallel grippers. Even simple grippers typically require grasp detection and the prediction of the grasp pose to handle unknown objects using external vision \cite{tian2022data}. However, objects are challenging to grasp in cluttered environments, and the environment constantly changes as the gripper interacts with it. To address this challenge, we developed a gripper that can both grasp and manipulate small objects while the object is between the fingers.

Given the nature of cluttered environments, it is unrealistic to expect the object to be automatically centered between the gripper's fingers even after successfully grasping an object. For this reason, we integrated an additional DOF into the gripper to ensure a stable grasp.

\begin{figure}[t]
\vskip 0.1in
\begin{center}
\centerline{\includegraphics[width=\columnwidth]{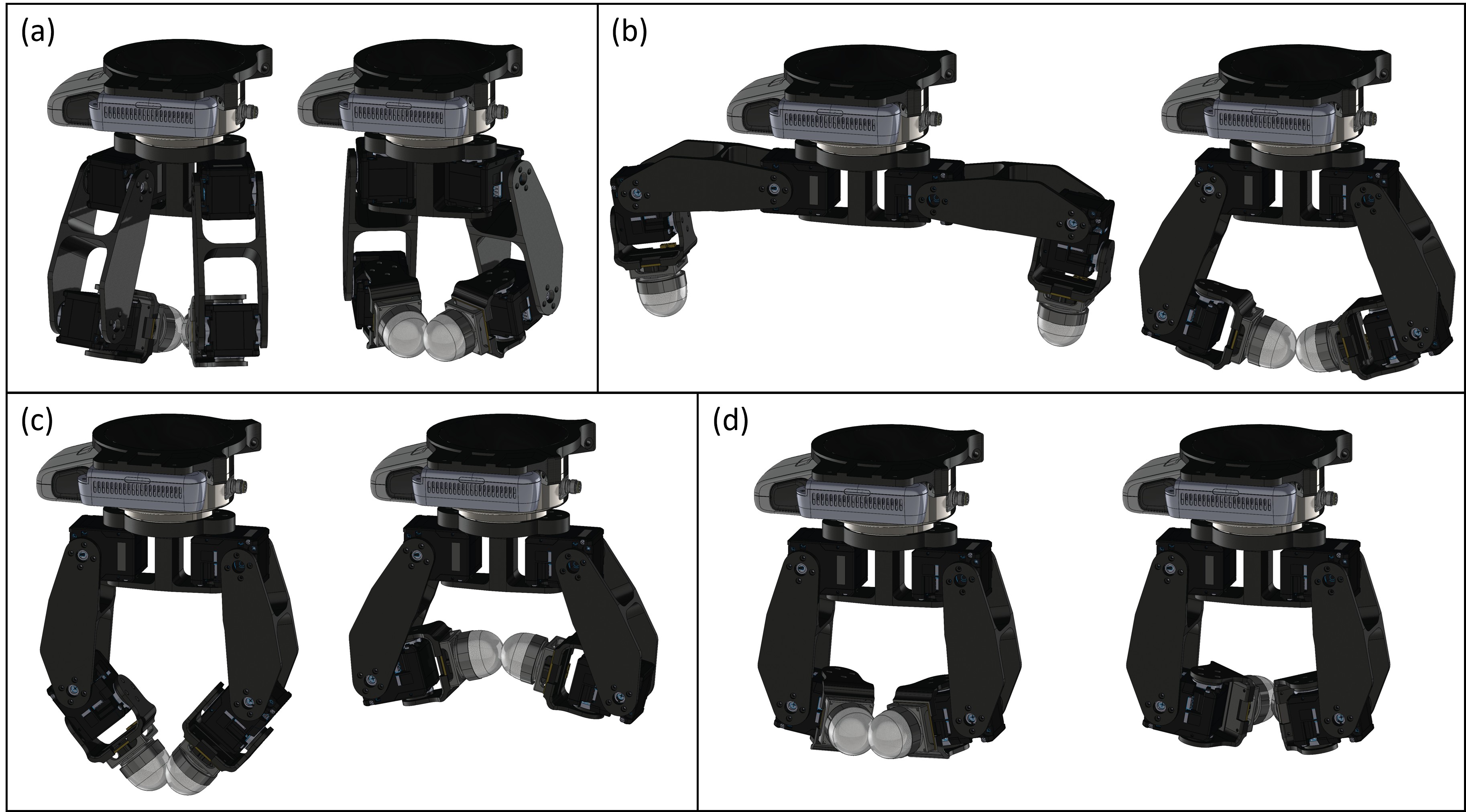}}
\caption{Joint limit of the gripper for each joint. }
\label{pic:gripper_range}
\end{center}
\vskip -0.4in
\end{figure}

We implemented four degrees of freedom for each finger to maximize inter-finger manipulation during grasping, enhancing the manipulation range during successful grasps. Effective inter-finger manipulation demands a maximal contact area between the fingertips of each finger. Assuming a deformable, hemispherical fingertip shape—beneficial for the unpredictability of cluttered environments—the contact workspace can be maximized if we rotate the fingertip in multiple directions while maintaining contact, ensuring the object remains securely held.
\begin{figure}[t]
\vskip 0.1in
\begin{center}
\centerline{\includegraphics[width=\columnwidth]{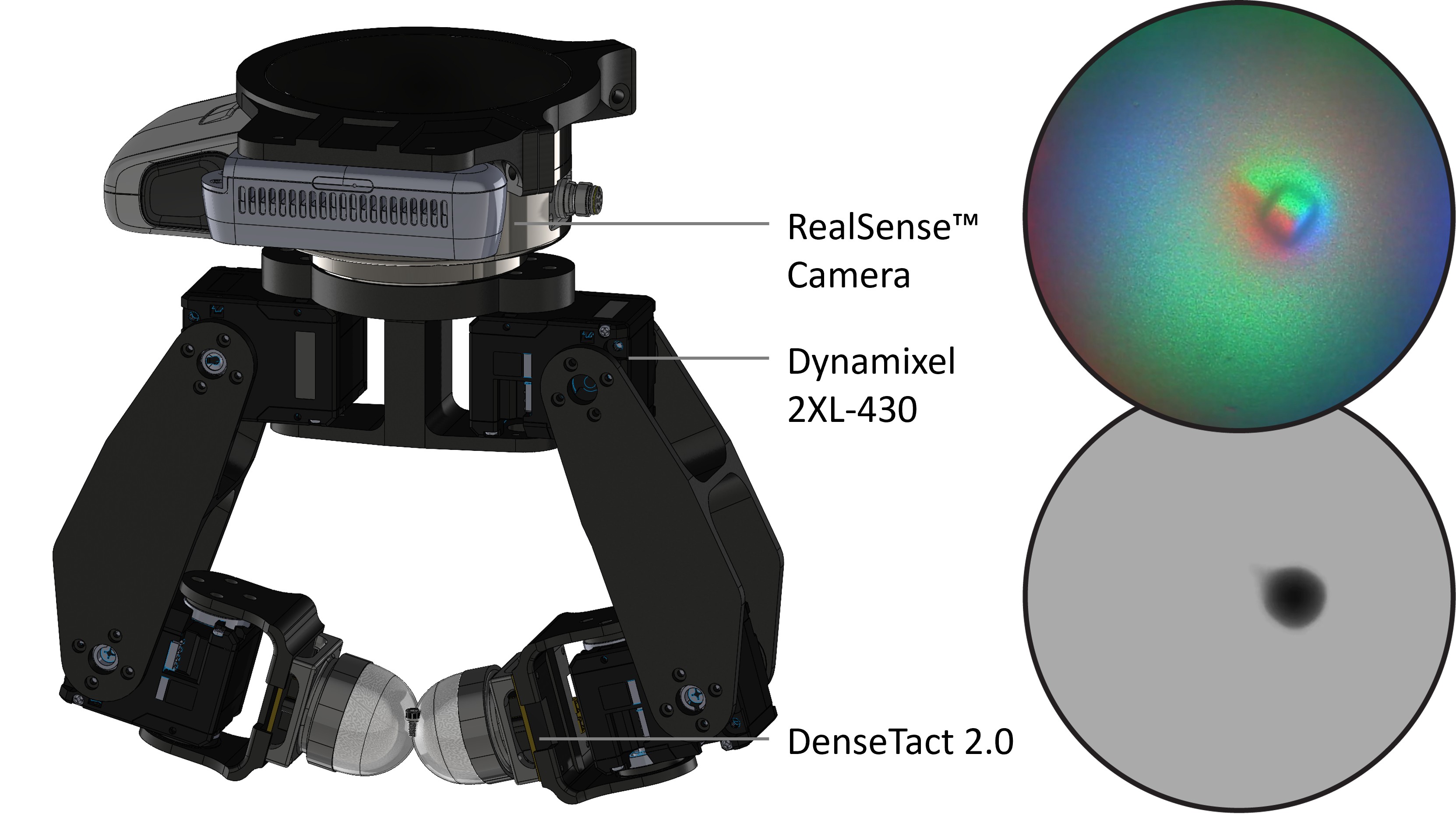}}
\caption{8-DOF gripper realistic model. The gripper can be attached directly to the Franka robot arm.}
\label{pic:gripper}
\end{center}
\vskip -0.4in
\end{figure}
To meet these requirements, we introduce a gripper design featuring two fingers with 4  DOF revolute joints for each finger. The left side of Fig. \ref{pic:gripper} presents this design, accompanied by a camera. As the fingers make contact and grasp, two R joints in each finger control the rotational movements in the x-direction and another two R joints for z-directions in the gripper frame, as shown in Fig. \ref{pic:ctrljnt}. These additional degrees of freedom offer enhanced control during the grasping of small objects. The working range of the gripper while making contact between two fingers is shown in Fig. \ref{pic:gripper_range}. 

The gripper uses four 2XL430-W250-T Dynamixel motors, boasting a total of eight degrees of freedom. This provides the gripper with a more expansive workspace and allows for dexterity beyond that of a conventional two-finger gripper. The gripper's arms are 3D-printed, ensuring it remains lightweight and reduces load.

\subsubsection{Dimensions}

The gripper is installable to the Franka robot arm by replacing the end-effector. The size of the arm between each joint is $(p_{12},\, p_{23},\, p_{34},\, p_{4ft}) = (24mm,  \quad 95.52mm, \quad 24mm, \quad 55mm)$, where $p_{ij}$ is the length between i-th and j-th joint, where $ft$ refers to the fingertip. The diameter of the hemispherical part of the fingertip is 31mm, which is suitable for grasping a small object with a size of $5mm\ \sim \,25mm$. STL and URDF files, complete with accurate mass and inertia values, are available on the project webpage.

\subsubsection{Tactile compliance design}

Soft fingertips in grippers have been shown to facilitate in-hand manipulation \cite{ciocarlie2007soft, lu2019soft}. We chose the Densetact 2.0 sensor as the gripper's fingertip due to its compliant gel component \cite{do2022densetact, do2023densetact}, which offers advantages over flat-surface sensors like Gelsight and Digit \cite{yuan2017gelsight, lambeta2020digit}. The soft gel enhances the contact area and friction, leading to secure grasps, especially for small objects.

The compliant nature of the Densetact 2.0 gel allows the gripper to adapt to uncertainties and distribute force more evenly during grasping. This adaptability is particularly useful for handling objects of varying shapes and poses. In contrast to sensors like SoftBubble \cite{alspach2019softbubble}, Densetact 2.0's hemispherical design offers a larger sensing area per volume, contributing to more precise in-hand manipulation.

When integrated with our multi-DOF gripper, the compliant features of the Densetact 2.0 silicone enhance the gripper's versatility for manipulating a diverse range of objects. The fingertip can deform up to 20mm, facilitating a secure grasp and minimizing object damage. Additionally, Densetact 2.0's deformation feedback aids in precise control, crucial for tasks like object orientation and dense-environment grasping.


\subsection{Perception from Tactile Sensor} \label{sec:perception}

We used a tactile sensor on the gripper's end-effector to determine the object's pose and position. Opting for a pattern-less DenseTact 2.0 sensor, we focused on sensor deformation instead of force estimation, given the object's negligible mass and our quasi-static manipulation assumption. The sensor was calibrated using the method in \cite{do2023densetact}, enabling depth image-based point cloud generation.

\begin{figure}[t]
\vskip 0.1in
\begin{center}
\centerline{\includegraphics[width=\columnwidth]{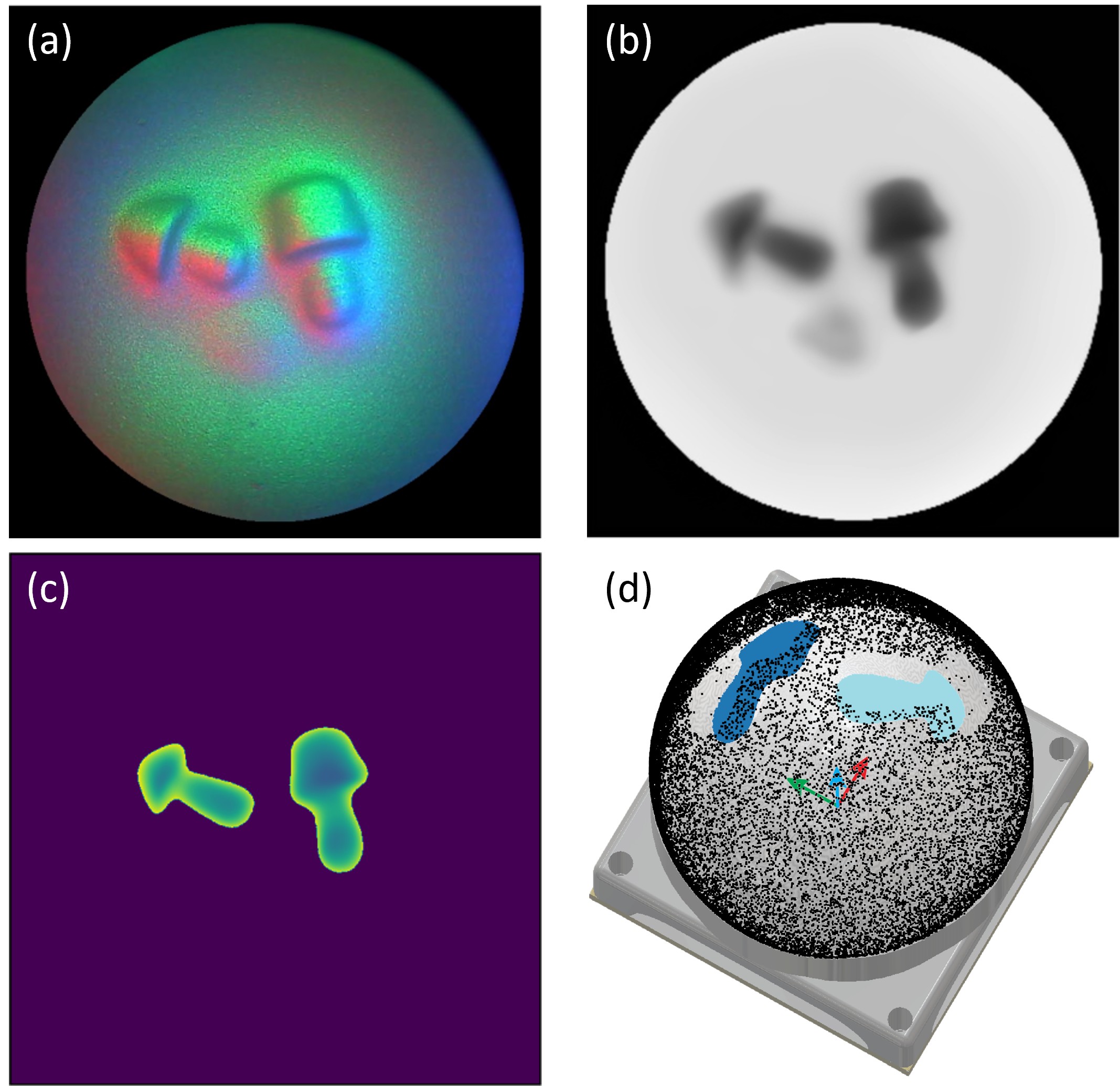}}
\caption{The tactile sensor measurement results are as follows: (a) displays the sensor's captured RGB input, while (b) presents the estimated depth output. (c) features the filtered and labeled depth output, and (d) illustrates the clustered point cloud using DBSCAN, overlaid on the tactile sensor.}
\label{pic:perception}
\end{center}
\vskip -0.4in
\end{figure}

For experiments, we isolated relevant points from the point cloud by setting a 3mm threshold against the undeformed state. We then segmented the deformed points using DBSCAN \cite{ester1996density}, with specific distance and sample count parameters. A random 4\% sample from the undeformed section was added to improve clustering. DBSCAN was chosen for its real-time applicability over alternatives like HDBSCAN \cite{mcinnes2017hdbscan}. During real-time control, the point cloud was truncated to 5000 points, allowing a frame rate of $10 \sim 13\text{Hz}$ on an Intel Core i7-11800H CPU. Fig. \ref{pic:perception} shows the segmented point cloud.

After segmenting, up to four labels were extracted, allowing the sensor to recognize a maximum of four objects per perception step. The label with the most points was prioritized during control. For classification, all labels contributed to the training dataset. Thus, the labeled point cloud is effectively transformed into inputs for either control or classification tasks.
\subsection{Grasping small object in cluttered environment}

Grasping a small object in a cluttered, ever-changing environment is challenging. We tackled this issue by using depth data from the robot arm and the adaptive capabilities of soft fingertips. The use of soft fingertips enhances adaptive grasping capabilities, especially in cluttered environments characterized by high uncertainty. Thus, if the gripper can position itself to the desired point in the cluttered environment, a simple closing motion with the soft fingertip can easily achieve grasping, even in a highly uncertain cluttered environment.

However, objects need to be in a specified region of interest for successful grasping. The challenge amplifies in a cluttered space with diverse small objects, like a bowl filled with assorted items, as the pile's profile changes during grasping attempts. To counter this, we use depth data from a RealSense camera on the Franka arm to identify the highest points in our target area, shown in Fig. \ref{pic:gripper}.

Our goal is to fine-tune the gripper's pose to maximize grasping success. We determined the optimal position and orientation within a square region, 24mm on each side, centered on the target point. From the camera's depth information, we extracted the top 800 elevation points in our target region. We then calculated the average position of these points in the world frame, represented as \({}^W\mathbf{p}_{\text{mean}} = [x_{\text{mean}}, y_{\text{mean}}, z_{\text{mean}}]^T\). A vector, \(\mathbf{v}\), is defined as the difference between this mean position and the center of the opening rim of the bowl (fixed point), \({}^W\mathbf{p}_{\text{cen}}\). Additionally, the orientation angle, \(\theta\), the angle for the 7th joint of the franka arm, is derived from the horizontal components of this vector:

\begin{align}
{}^W\mathbf{v} &= {}^W\mathbf{p}_{\text{mean}} - {}^W\mathbf{p}_{\text{cen}}, \quad \theta = \tan^{-1}\left(\frac{v_x}{v_y}\right)
\end{align}


During the Detection and Grasping phase, the gripper first moves to the position 60cm above the desk. As shown in the first bottom left image of Fig. \ref{pic:pipeline}, the depth camera detects the pile and returns the position to grasp. During this stage, the gripper remains open. Next, the gripper moves to the center position, adjusting the orientation of the last joint by the computed rotation angle, \(\theta\). Following this adjustment, the gripper advances guided by the vector \(\mathbf{v}\), ensuring its trajectory towards the pile is both optimal in angle and position, thereby maximizing the success rate of the grasp. Finally, the gripper grasps the object by closing the gripper, and we move the gripper's position 2 seconds after the gripper grasps the object.

\subsection{In-hand Orientation of Small Objects}

After the gripper grasps an object and detects it within the fingertip via DenseTact, the small objects that have been grasped often deviate from the center of the gripper's fingertip. This deviation necessitates additional inter-finger manipulation for a stable grasp and proper classification. To address this challenge, we introduce a control strategy for securely grasping unknown small objects utilizing tactile feedback.

Even though the initial state of the gripper remains consistent, the objects it grasps are unpredictable and unfamiliar. Consequently, the controller's primary goal is to align the fingertip's position with the detected object while ensuring consistent pressure between the two fingertips of the gripper. Therefore, the objectives of our controller are to: 1) to maintain a specific distance between the fingertips, ensuring a stable grasp, 2) to maneuver the gripper within its joint limits; and 3) to center the fingertip's origin with the grasped object.

We select the state of our controller as 
\begin{align}
\mathbf{x} = \{{}^Gy, {}^G\theta_x, {}^G\theta_z\} \in \mathbb{R}^3    
\end{align}
where the ${}^G$ refers to the gripper frame, ${}^Gy$ is the y-coordinate position of the fingertip in the gripper frame, ${}^G\theta_x, {}^G\theta_z$ are the angles of the fingertip coordinate frame in x and z axis of the gripper frame respectively, as defined in the left image of the Fig. \ref{pic:ctrljnt}. According to the figure, the Jacobian of one finger can be defined as the following: 

\begin{align}
\dot{\mathbf{x}}_{all} = J_{all} \,\, \dot{\mathbf{q}}, \quad J_{all} = \begin{pmatrix}J_v & J_w \end{pmatrix}^T, \quad J_{all}\in \mathbb{R}^{6 \times 4}
\end{align}

Where $\mathbf{x}_{all} \in \mathbb{R}^6$ refers to the position and angular position of the fingertip, and $\mathbf{q} = \begin{pmatrix}
    q_1 & q_2 &q_3 &q_4
\end{pmatrix} \in \mathbb{R}^4$ is the joint value of the one finger of the gripper. From the Jacobian of the joint, we can extract the corresponding differential value of each state by building the new Jacobian. Furthermore, since we have additional DOF for the new Jacobian, we can control the gripper to move within the joint limit through null space:

\begin{align}
    J = \begin{pmatrix}
        J_{v,2}\\ J_{w,1} \\ J_{w,3}
    \end{pmatrix} \in \mathbb{R}^{3 \times 4}, \quad J^{\dagger} = (J^T J + \lambda^2 I)^{-1} J^T
\end{align}

Where $J_{v,i}$ or $J_{w,i}$ is the i-th row of the velocity Jacobian or angular Jacobian respectively.  Then, the desired joint position can be computed by integrating the desired velocity through a geometric controller. The desired joint velocity can be computed as: 
\begin{figure}[t]
\vskip 0.1in
\begin{center}
\centerline{\includegraphics[width=\columnwidth]{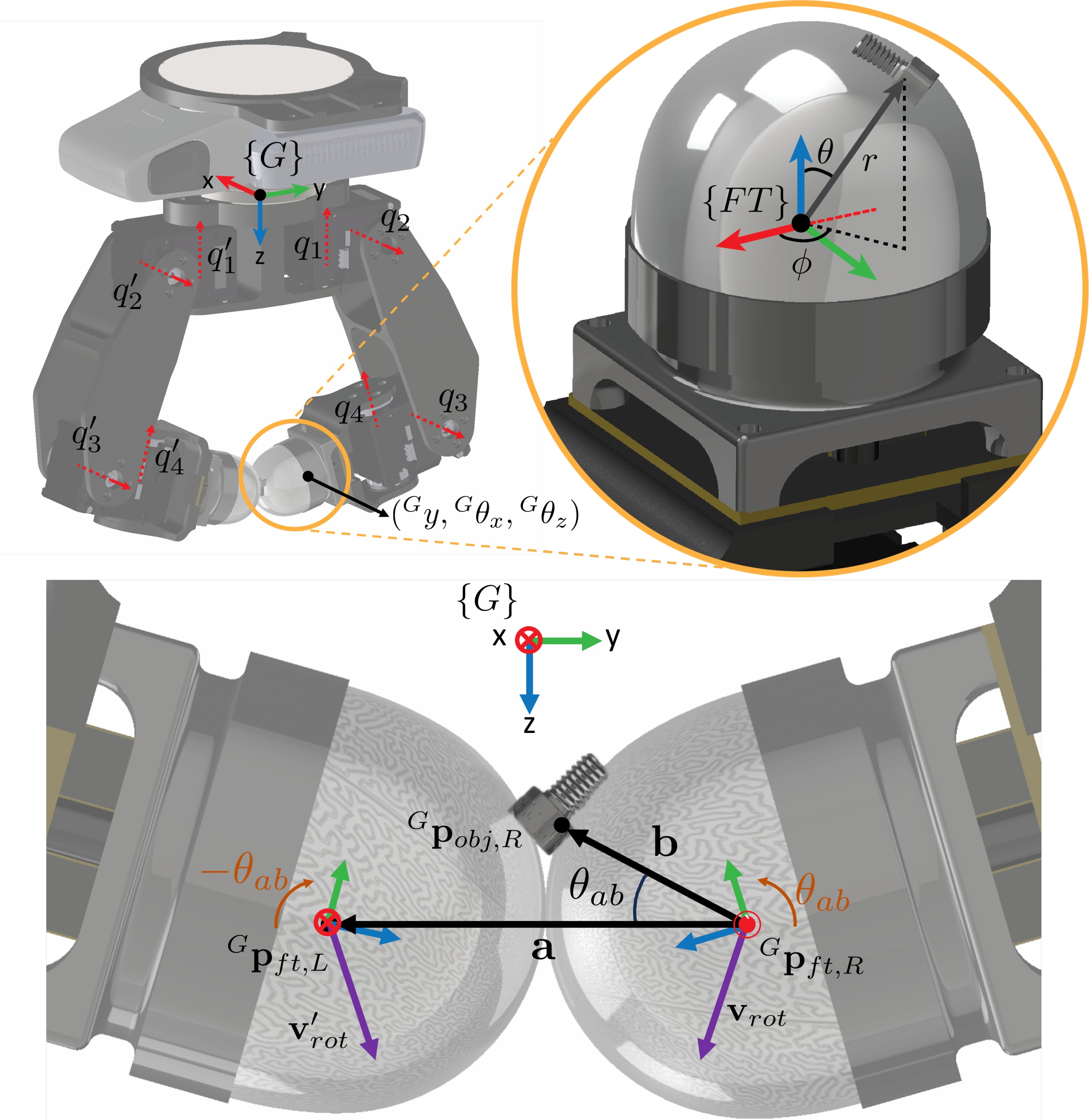}}
\caption{Axis of the gripper and controller input specified in the fingertip coordinate frame. The image below shows the magnified view of the gripper while grasping an object.}
\label{pic:ctrljnt}
\end{center}
\vskip -0.4in
\end{figure}

\begin{align}
\dot{\mathbf{q}} = J^{\dagger} \mathbf{v}_{\text{des}} + (I - J^{\dagger} J) f_{pen}(\mathbf{q})
\label{eqn:main_control}
\end{align}

Where $f_{pen}(\mathbf{q}) =  -C(\mathbf{q}_{curr}- \mathbf{q}_{mid})$. 
$f_{pen}$ refers to the penalty term to ensure the joint inside of the range, $C$ is constant for the penalty term, $\mathbf{q}_{mid}$ is the median of joint value in the joint range. The $v_{des}$ is the desired fingertip velocity. From the desired fingertip velocity, we can get the desired joint position command. 

Due to the absence of prior information about the object, and given our objective is its classification, the controller's goal needs dynamic adjustments. Based on tactile sensor input, we compute the controller's goal as \textbf{minimizing the $\theta_{ab}$ while maintaining the grasp of object}, where $\theta_{ab}$ is defined as 
\begin{align}
\theta_{ab} &= \cos^{-1}\frac{\mathbf{a} \cdot \mathbf{b}}{|a| |b|}, \quad \hat{\mathbf{v}}_{rot} = \frac{\mathbf{b}\times \mathbf{a}}{|a| |b|} \\
\mathbf{a} &= {}^Gp_{ft,L} - {}^Gp_{ft,R},\quad \mathbf{b} = {}^Gp_{obj,R} - {}^Gp_{ft,R}
\end{align}

Where ${}^Gp_{ft,L}$ and $ {}^Gp_{ft,L},$ are the position of the left and right fingertip (center of the hemispherical tactile sensor, the origin of the fingertip coordinate frame), and $ {}^Gp_{obj,R}$ is the position of the detected object in the right fingertip in the gripper frame. Those values are also shown in the bottom image of Fig. \ref{pic:ctrljnt}. We derive $ {}^Gp_{obj,R}$ by averaging the bottom 30\% of point clouds with the lowest deformation values. This means we leverage the point clouds that occupy the top 30\% in terms of the $\mathbf{r}$ value, as illustrated in the top right image of Fig. \ref{pic:ctrljnt}. This strategy lets the gripper determine the subsequent movement point without settling on the currently detected state.
 From the above value, we can get the desired velocity: 
\begin{align}
    v_{des} = K_p\begin{pmatrix}
       \dot{\mathbf{p}}_{ft,R}\\
       \omega_x  \\
       \omega_z 
    \end{pmatrix} \quad  = \frac{K_p}{\Delta t}  \begin{pmatrix}
       C_y - \lambda \\
        \hat{\mathbf{v}}_{rot,x} {\theta_{ab}} \\
        \hat{\mathbf{v}}_{rot,z} {\theta_{ab}}
    \end{pmatrix}
\end{align}
Where  $K_p \in \mathbb{R}^{3\times 3}$, $\Delta t$ is a constant value, $C_y$ is a constant value which refers to the offset of the fingertip from the contact, and $\lambda$ is the deformed radius value of the detected object in the DenseTact. From the first row, the gripper can maintain constant pressure while keeping contact between the fingertip and the object. Since the gripper detects the object before the controller starts, the object will always exist while the controller is executed. $\hat{\mathbf{v}}_{rot,x}, \hat{\mathbf{v}}_{rot,z}$ are the x and z components of the rotation axis, respectively. The process finishes when the fingertip and the object's center align. The controller is operated for one finger of the gripper, and the other finger gets the same value to achieve the symmetric movement for a stable grasp rolling without slipping.

Given the small, lightweight nature of the target objects, inertial and force inputs are less relevant and unpredictable. We thus use a position controller that integrates the commanded velocity (Eqn. \ref{eqn:main_control}) in a quasi-static state, and this controller choice is driven by motor control limitations as well as the pipeline's real-time processing speed (10Hz \~ 13Hz).

While the controller could be modeled through optimization or RL, these options present challenges. The complex gel deformation we're tackling is best represented by hyperelastic material models like the Ogden hyperelastic model, which require computationally heavy FEM programs \cite{ogden1972large}. Additionally, RL or dynamic learning approaches often need extensive simulated data, making them less practical for our task. Other issues involve sim-to-real gaps and errors in dynamic modeling.

\begin{figure}[t]
\vskip 0.1in
\begin{center}
\centerline{\includegraphics[width=\columnwidth]{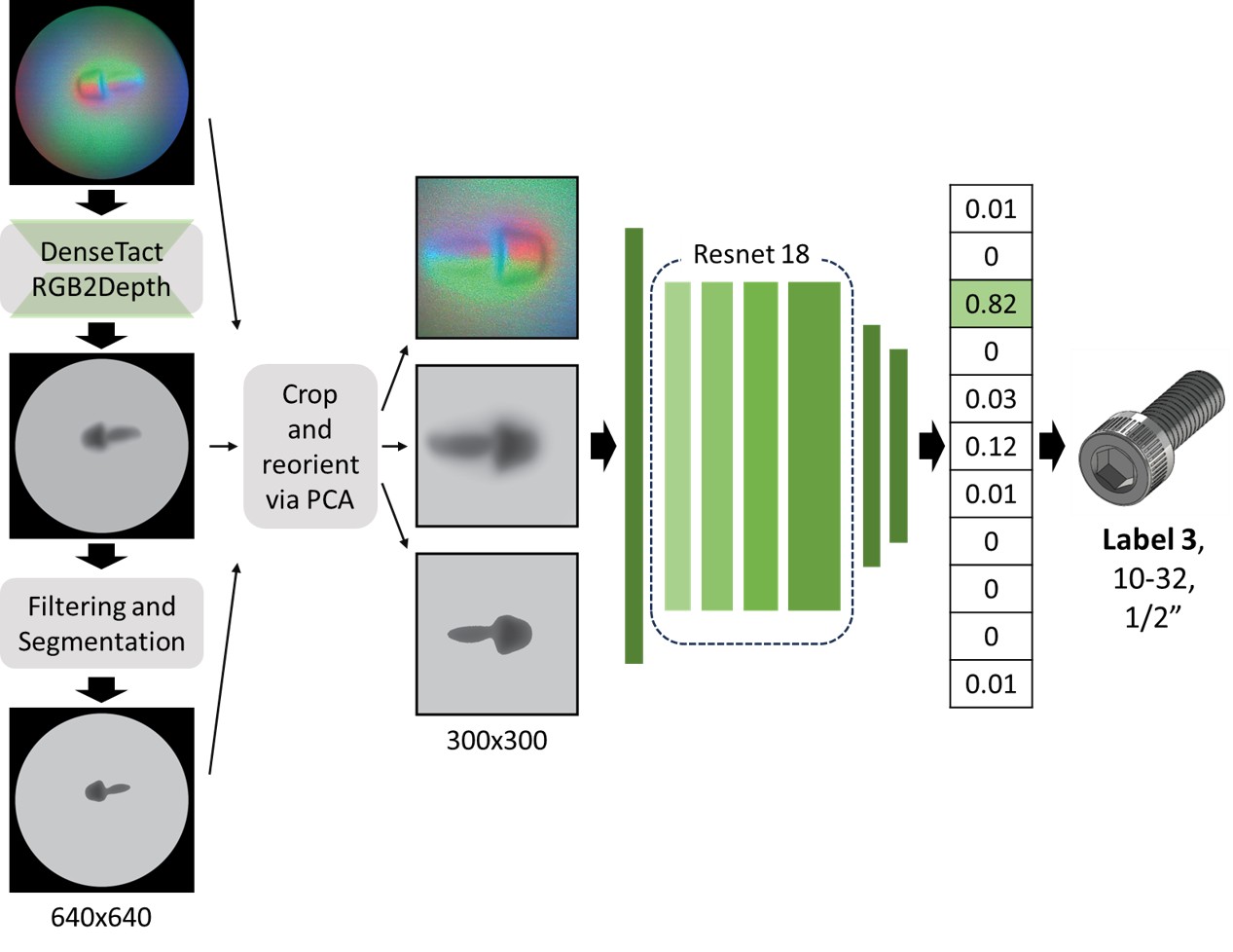}}
\caption{Pipeline architecture for small object classification.}
\label{pic:network}
\end{center}
\vskip -0.3in
\end{figure}
\subsection{Small Object Segmentation}

\subsubsection{Dataset Collection}

Tactile sensors are crucial for object identification in grasping, particularly with soft fingertips that significantly occlude the object. External vision proves insufficient for object verification in such cases. Leveraging the high-resolution ($640 \times 640 \times 3$) input from tactile sensors, we curated a dataset of objects grasped between two such sensors. Each object was positioned on one sensor and encapsulated by pressing the other sensor onto it. Live RGB and depth images were captured once the object became discernible. For each object type, 50 RGB and 50 corresponding depth images were collected during a single press.

The dataset also accommodates scenarios of grasping one or two small objects simultaneously. We focused on select combinations due to the exponential increase in potential object pairings, denoted by $\frac{n(n-1)}{2}$ for $n$ different objects. Our dataset comprises 20 types of small objects, including 9 varieties of screws and 11 daily objects, most of which are easily graspable by the gripper.

The screw dataset was designed such that the screws vary by length, head diameter, and thread size. Different combinations of variables are changed across the dataset classes. Each screw is either 1/2'', 3/8'', or 1/4'' in length. Furthermore, each screw has a head diameter and thread spacing combination of either 4-40, 4-48, 10-24, 10-32, or 1/4''-28.

\subsubsection{Preprocessing step of the Image Input}

Given the relatively small size of the dataset when compared to the variety of object types (20 distinct objects), directly utilizing the raw input from the tactile sensor becomes impractical. Additionally, there's a potential for classification errors when the gripper unintentionally captures two small objects simultaneously. This challenge can be addressed by integrating an additional input layer and conducting suitable image preprocessing.

Initially, we incorporated input from the labeled image derived through DBSCAN, along with the RGB and depth images generated by the tactile sensor. Following this, the labeled and deformed pointcloud was extracted and projected onto the depth image. We then employed PCA analysis to ascertain the orientation and center of the deformed point. Given the prior labeling of the deformed pointcloud, PCA analysis was conducted for each labeled pointcloud. As indicated in section \ref{sec:perception}, PCA can handle up to four labels in a single tactile input.

Relying on the central values and angles obtained from the PCA, the images were cropped to a size of $300 \times 300$, and rotated according to the identified angle. By integrating RGB, depth, and labeled images, the resulting input dimension became $300 \times 300\times 5$. This preprocessing approach enhances the efficiency of network training, even with a limited dataset size. Furthermore, the labeling step allows the localization of the classified objects and completes the segmentation of the multiple objects detected from the raw sensor image.

\subsubsection{Model for Classification}

The network architecture chosen for classification is grounded in the ResNet18 framework, a decision driven by the compact size of our dataset, as shown in Fig. \ref{pic:network}. Rather than maintaining the model in its static form and solely training the concluding MLP layer, we opted to activate the final fourth layer block for training while keeping the other layer blocks of ResNet18 static \cite{he2016deep}. Preceding the initial layer block of ResNet18, a 2D convolutional layer accepts the input, which is subsequently processed through batch normalization, ReLU, and a max-pooling layer. After the fourth layer block of ResNet18, two fully connected layers are employed, utilizing a hidden channel size 256. Ultimately, a softmax function is invoked to classify the object type. 

Training was conducted on a composite dataset, incorporating single-object and multi-object datasets. This amalgamation inherently led to a disparity in the dataset count for individual object types. To counterbalance this, 12\% of the total dataset was randomly collected as a testing set for every object category. The number of datasets per class was recorded while splitting training and testing datasets. This count was then employed as a weight in the cross-entropy loss calculation throughout the training phase. Utilizing the Adam optimizer, we set a learning rate of $2\times 10^{-5}$ and a weight decay of $1\times 10^{-4}$ over a span of 400 epochs and with a batch size of 8. The training duration was approximately an hour, executed on four NVIDIA A4000 GPUs.



\section{Experiment}\label{sec:experiment}

\begin{figure}[t]
\vskip 0.1in
\begin{center}
\centerline{\includegraphics[width=\columnwidth]{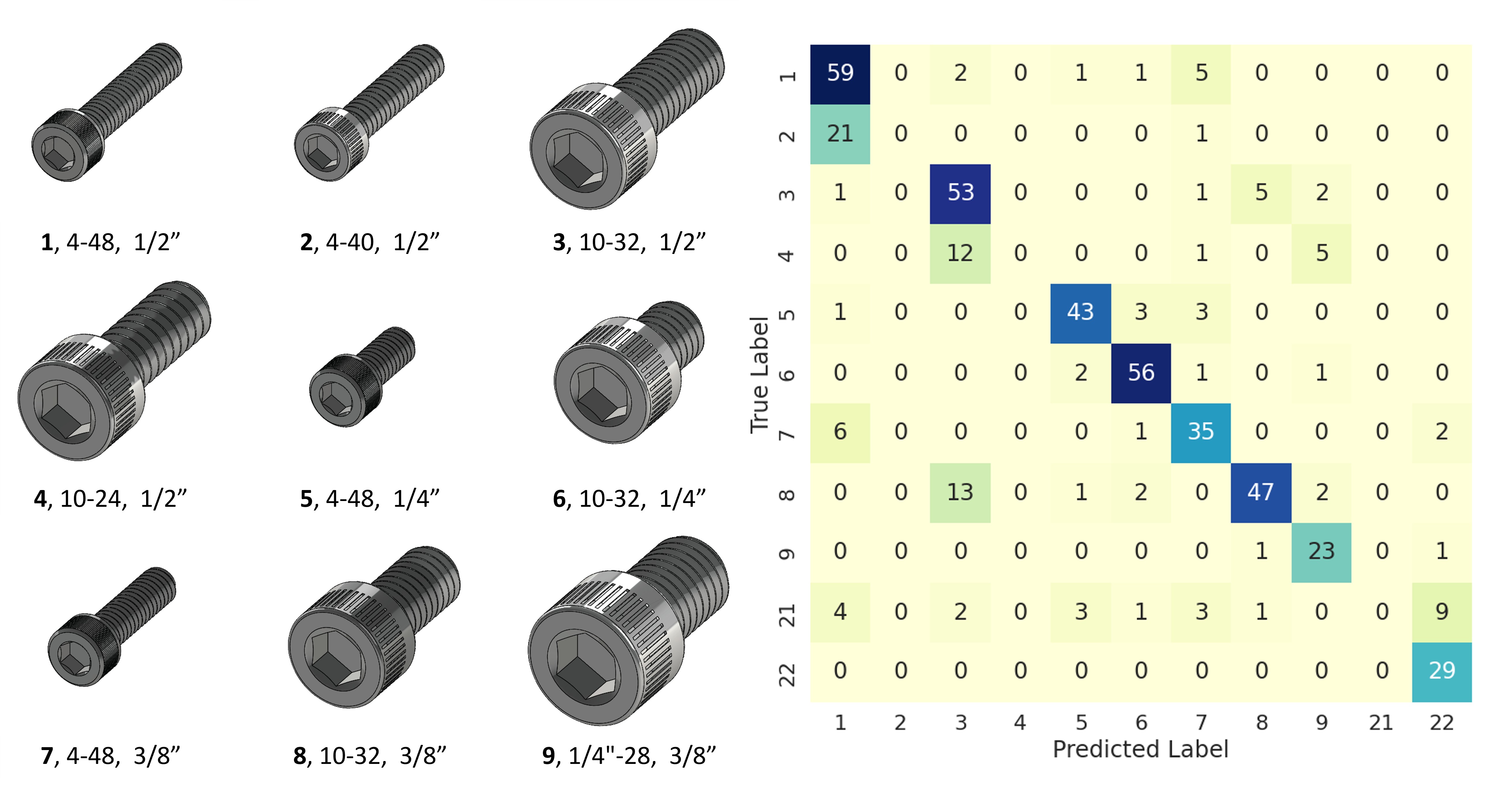}}
\caption{Classification result of screw objects and classified labels for 466 touches.}
\label{pic:conf_screw}
\end{center}
\vskip -0.45in
\end{figure}

\subsection{Classification}

Before physically demonstrating the complete procedure, the classification results were evaluated using the test dataset. The right image in Fig. \ref{pic:conf_screw} displays the confusion matrix for the classified screws. In the left image, the classified label, thread size, and length of each screw are indicated at the bottom of their respective images. Label 21 is designated for instances involving two screws, while Label 22 signifies that the sensor either detected a plane or failed to detect the screw. Given that Labels 1 and 2 share identical lengths and head sizes but differ in thread type, it's understandable that Label 2 is occasionally classified as Label 1. A similar misclassification occurs between Labels 3 and 4. Due to the combinations required for a two-screw dataset exceeding 45 distinct cases, achieving uniform dataset size via human input proved challenging. Nonetheless, by accumulating a broader combination of datasets or gathering additional data during demonstrations, we believe classification errors can be reduced.

\begin{figure}[t]
\vskip 0.1in
\begin{center}
\centerline{\includegraphics[width=\columnwidth]{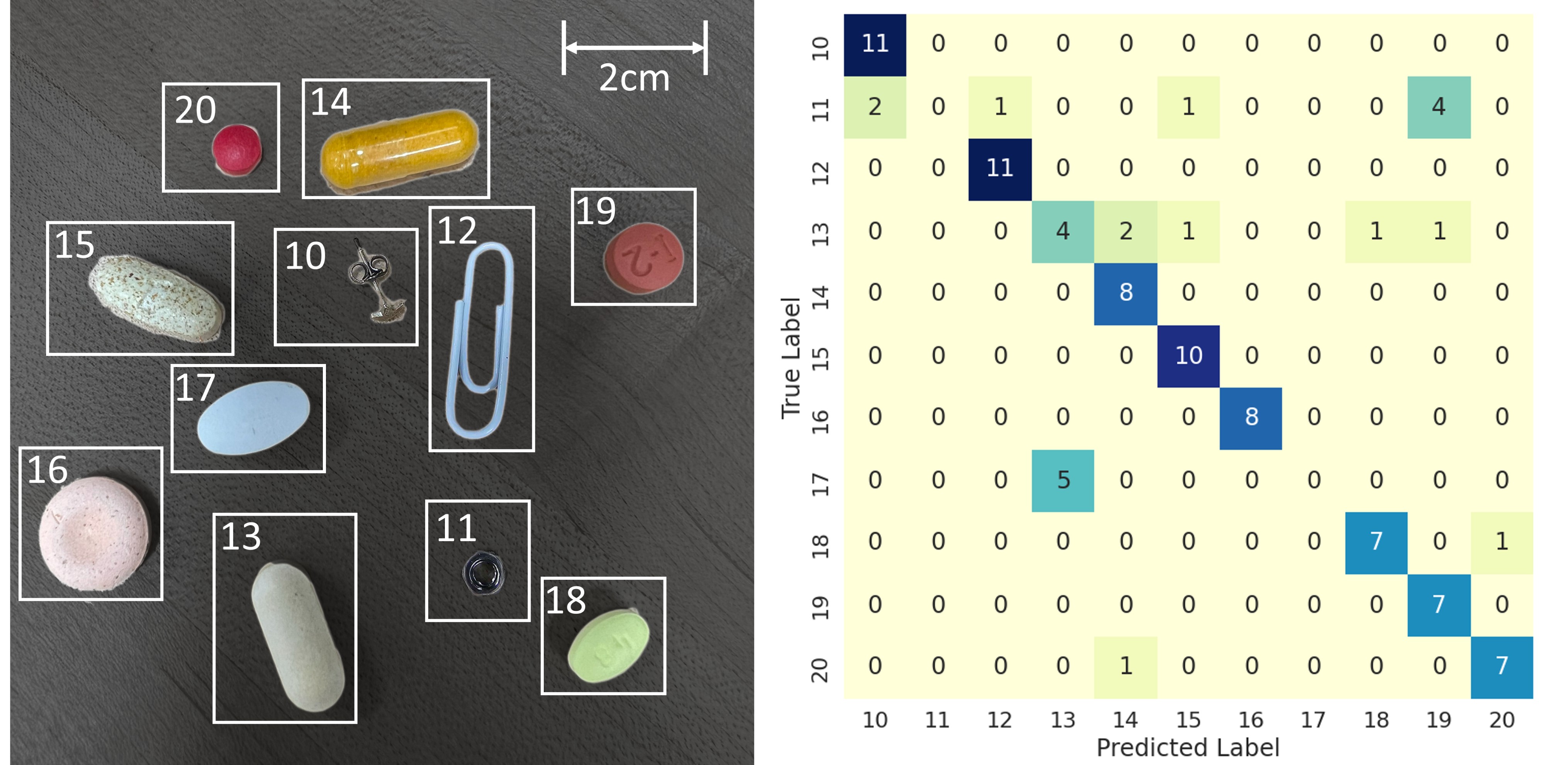}}
\caption{Classification result of small random objects and classified labels for 93 test touches.}
\label{pic:conf_diffobj}
\end{center}
\vskip -0.45in
\end{figure}

Fig. \ref{pic:conf_diffobj} presents the confusion matrix for the classification of other objects. The left image annotates the corresponding labels for each object. We opted for a diverse set, encompassing various types of pills, earrings, paperclips, and more. Despite the dataset's limited size, the classifier has demonstrated proficiency in correctly identifying most objects. 
\subsection{Full pipeline}

Building on the methods described and as shown in Fig. \ref{pic:pipeline}, we structured the grasping, reorienting, and classification sequence into a finite state machine with several key states: `initial,' `detect,' `ready for grasp,' `grasp,' `control,' and `classification.'

In the `detect' phase, the system cycles back to detection if the depth camera provides inadequate sensor values. Sensor feedback determines grasp success after the `grasp' phase; failure redirects the process back to the `detect' phase. During `classification,' if the sensor identifies `two screws' or `plane,' the system returns to the `initial' state for a new cycle.

Given the small object sizes and the sensor's high deformability, one tactile sensor usually suffices for object detection and manipulation. Experiments were conducted using output from a single sensor while the other finger moved in tandem. As the objects classified are symmetrical, consistent sensor readings are assured for both fingers. Classification of asymmetric objects, though feasible, would require data from both object facets.

\subsection{Demonstration result}

\begin{figure}[t]
\vskip 0.1in
\begin{center}
\centerline{\includegraphics[width=\columnwidth]{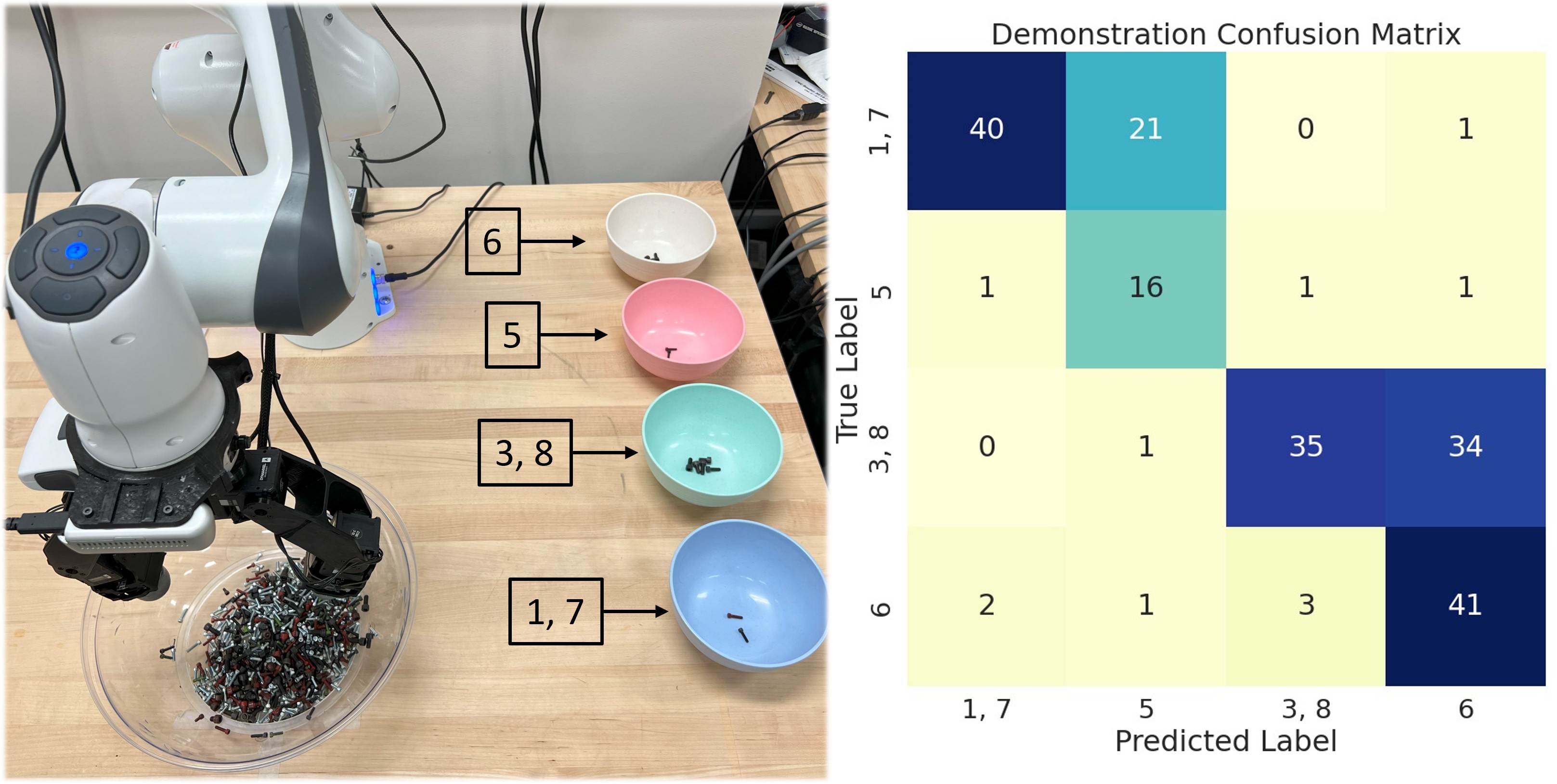}}
\caption{Experimental setup and demonstrated result of the whole process for 198 grasps.}
\label{pic:conf_demon}
\end{center}
\vskip -0.45in
\end{figure}

Utilizing the established pipeline and integrating all processes, we executed the object sorting task autonomously, eliminating the need for human intervention. The left side of Fig. \ref{pic:conf_demon} illustrates the experimental setup during the demonstration, whereas the right side depicts the confusion matrix derived from the demonstration results. The cluttered environment is represented in a transparent bowl and only the depth camera was employed to detect the highest point of the pile. For this experiment, all objects in the bowl are screws. The pile comprises 50 objects with Labels 3, 6, and 8. These objects are considerably larger in size compared to others. Additionally, 100 objects with Labels 1, 5, and 7, recognized by their smaller head size, are present within the environment. The variation in object numbers stems from the gripper's inherent tendency to seize larger objects, due to its grasping characteristics. Importantly, the entire grasping process remained devoid of human influence (for instance, altering the pile profile during the demonstration or manual re-grasping). The gripper still exhibited a marked preference for grasping larger-headed screws.

The process of grasping the objects proved largely successful. Out of 225 attempts, there were 198 successful grasps. In contrast, there were 12 instances of unsuccessful grasping and 12 trials where the results were classified under Labels 21 or 22, indicating scenarios where two screws were grasped simultaneously or when a plane was detected. Consequently, \textbf{88\%} of the trials resulted in successful object extraction from the cluttered environment and subsequent object classification.

The results presented in Fig. \ref{pic:conf_demon} highlight a recurring misclassification. Specifically, objects with Labels 1 or 7 were frequently mistaken for Label 5, while objects Labeled 3 and 8 were often misclassified under Label 6. This trend can be attributed to the gripper's occasional tendency to grasp the head of the screw first and hold the grasp. When this happens, even after the finger position is changed, there is a possibility that the sensor only observes the head part of the screw. Both Labels 1 and 7 possess long screws that share head sizes with Label 5, while Labels 3 and 8 have similar characteristics with Label 6. One of the results of the example can be observed in the sensing results displayed in Fig. \ref{pic:pipeline}. Given that objects under Labels 1, 5, and 7 have identical screw heads, misclassifications amongst them are plausible. However, Labels 5 and 6 are not mistakenly classified under other Labels, mainly due to the shorter lengths of these objects, which increased the likelihood of head detection during dataset collection. There were instances of failed trials where the gripper occasionally grasped two objects simultaneously, rendering the secondary object invisible to the sensor. Such challenges could be potentially addressed by leveraging sensor feedback from both fingertips.

\section{Conclusions}\label{sec:conclusion}
In this study, we present a novel approach for manipulating and classifying small objects in cluttered settings using optical tactile sensors. Our key innovation is a gripper fitted with DenseTact 2.0, designed to enhance both grasping success and post-grasp manipulation, thanks to its highly deformable soft fingertip. A unique manipulation strategy using a newly devised Jacobian combination ensures stable grasps and precise classification.

Our network model efficiently classifies objects, even with a limited dataset, demonstrating broad applicability to general small objects. The end-to-end pipeline operates autonomously, underscoring the potential for human-free small object classification and manipulation. This work not only advances current grasping strategies and object pose estimation techniques but also lays the groundwork for more versatile robotic grasping solutions.

Future research could focus on using tactile sensors on both fingertips for more stable grasping and enhanced classification, addressing the concurrent grasping of multiple objects, and extending the gripper's utility in human-robot collaborative settings.

\addtolength{\textheight}{-12cm}   



  \bibliographystyle{./IEEEtran} 
  \bibliography{./IEEEexample}

\end{document}